\documentclass[letterpaper]{article} % DO NOT CHANGE THIS
\usepackage{aaai24}  % DO NOT CHANGE THIS
\usepackage{times}  % DO NOT CHANGE THIS
\usepackage{helvet}  % DO NOT CHANGE THIS
\usepackage{courier}  % DO NOT CHANGE THIS
\usepackage[hyphens]{url}  % DO NOT CHANGE THIS
\usepackage{graphicx} % DO NOT CHANGE THIS
\usepackage{amsmath}
\usepackage{natbib}  % DO NOT CHANGE THIS AND DO NOT ADD ANY OPTIONS TO IT
\usepackage{caption} % DO NOT CHANGE THIS AND DO NOT ADD ANY OPTIONS TO IT
\usepackage{algorithm}
\usepackage{algorithmic}

\usepackage{newfloat}
\usepackage{listings}
\DeclareCaptionStyle{ruled}{labelfont=normalfont,labelsep=colon,strut=off} % DO NOT CHANGE THIS
\floatstyle{ruled}
\newfloat{listing}{tb}{lst}{}
\floatname{listing}{Listing}
\title{Confidence Is All You Need for MI Attacks (Student Abstract)}
\author{
    Abhishek Sinha, Himanshi Tibrewal, Mansi Gupta, Nikhar Waghela, Shivank Garg }

\affiliations{
    Vision and Language Group, Indian Institute of Technology Roorkee, Roorkee, Uttarakhand, India - 247667 \\
   a\_sinha@ma.iitr.ac.in , himanshi\_t@ce.iitr.ac.in , m\_gupta@ma.iitr.ac.in , n\_waghela@ma.iitr.ac.in , shivank\_g@mfs.iitr.ac.in   
}

\usepackage{bibentry}
\begin{document}

\maketitle 

\begin{abstract}
In this evolving era of machine learning security, membership inference attacks have emerged as a potent threat to the confidentiality of sensitive data. In this attack, adversaries aim to determine whether a particular point was used during the training of a target model. This paper proposes a new method to gauge a data point's membership in a model's training set. Instead of correlating loss with membership, as is traditionally done, we have leveraged the fact that training examples generally exhibit higher confidence values when classified into their actual class. During training, the model is essentially being 'fit' to the training data and might face particular difficulties in generalization to unseen data. This asymmetry leads to the model achieving higher confidence on the training data as it exploits the specific patterns and noise present in the training data. Our proposed approach leverages the confidence values generated by the machine-learning model. These confidence values provide a probabilistic measure of the model's certainty in its predictions and can further be used to infer the membership of a given data point. Additionally, we also introduce another variant of our method that allows us to carry out this attack without knowing the ground truth(true class) of a given data point, thus offering an edge over existing label-dependent attack methods.
\linebreak

\end{abstract}

\section{Introduction}{
 In recent years, using MI (Membership Inference) attacks to predict whether a specific training example was used as training data in a particular model has garnered significant momentum. Currently, approaches rely predominantly on loss-based techniques, rooted in the fundamental concept that data points in the training set exhibit lower loss when processed by the target model. In \cite{shokri2017membership}, various strategies for executing MI attacks from first principles are outlined. The LOSS attack \cite{yeom2018privacy}, which initially showed promise with an accuracy of 60\%, fails to identify members effectively in CIFAR-10, rendering it impractical for membership inference. Following this, the Likelihood Ratio Test, \cite{carlini2022membership}, was introduced, improving the attack's success across various metrics, including accuracy and AUC. This method essentially computes the loss on a given data point and measures the likelihood of this loss under two worlds, one where a model is trained on this data point and the other where it is not.  
 \newline
 Building upon this concept, we propose using confidence values and measuring the likelihood of the confidence values under the distribution of models containing a given data point against the distribution of models that have never seen the given data point. Observing that the training set samples should ideally display a higher output confidence level is crucial. This compels us to investigate an alternative classification method based on confidence levels.This strategy is based on the premise that training examples should exhibit a specific level of confidence, and deviations from this pattern may indicate group membership or non-membership, as the model, through training, is learning the characteristics of the training data. In addition, traditional membership inference attacks rely on knowledge of the actual class label(ground truth) of the data point whose membership we infer. We investigate the possibility of removing the requirement of authentic identifiers from our classification strategy. This can be achieved by selecting the class with the highest predicted probability, effectively indicating the model's most confident prediction for the given input data point. This can be used as a proxy for the confidence value, and further, the attack can be carried out in the same way as before.  

}

\begin{table*}
  \centering
  \small
  \begin{tabular}{|c|c|c|c|c|c|}
    \hline
    & \textbf{Loss Value (Baseline)} & \textbf{Confidence Values} & \textbf{log(Confidence Values)} & \textbf{Argmax} & \textbf{log(Argmax)} \\
   
    \hline
    Attack(Online) & 0.5753 & 0.5668 & 0.575 & 0.5464 & 0.5447 \\
    \hline
    Attack(Online,FV) & 0.5879 & 0.593 & 0.6009 & 0.5622 & 0.5602 \\
    \hline
    Attack(Offline) & 0.5181 & 0.492 & 0.4721 & 0.478 & 0.4756 \\
    \hline
    Attack (Offline,FV) & 0.5184 & 0.4928 & 0.4804 & 0.4834 & 0.4815 \\
    \hline
    Attack Global Threshold & 0.5448 & 0.5439 & 0.5469 & 0.5376 & 0.5377 \\
    \hline
  \end{tabular}
  \caption{FV stands for Fixed Variance,Using AUC metric}
  \label{table1}
 
\end{table*}
\section{Methodology}{
 Drawing reference from \cite{carlini2022membership}, we commenced our process by training 16 shadow models, for 21 epochs each, on random samples drawn from the encompassing data distribution $D$. Precisely half of these models were trained on a target point $(x, y)$, whose membership or non-membership we sought to establish. The remaining models were trained without involvement with this particular point (These are the IN and OUT models, respectively). Then, we fit two Gaussian distributions to the confidence of the OUT and IN models on $(x, y)$ ( $Q\stackrel{\smash{\raisebox{-1ex}{\kern-0.75ex\scriptsize out}}}{}$ and $Q\stackrel{\smash{\raisebox{-1ex}{\kern-0.75ex\scriptsize in}}}{}$). While earlier the assessment of ‘confidence’ was accomplished by employing a logit scaled loss function (baseline), we used a more direct and intuitive method that aligns more closely with the inherent properties of a model’s prediction. Both of our proposed methods are based on the intuition that training examples exhibit higher output confidence scores than out-of-training data points. If we let $f(x)$ denote the evaluation of the model along with a final softmax activation function, we initially used the confidence probability associated with the actual class of the sample and studied the results. It is important to note that f(x) is a c-dimensional column vector where c denotes the number of classes and $y_{\text{true}}$ refers to the actual class of the target point.
\begin{equation}
\text{confidence} = \log\left(c = y_{\text{true}} (f(x)) + 1 \times 10^{-45}\right)
\end{equation}
This is logically sound as a higher confidence value calculated using our proposed metric, indicates a higher likelihood of belonging to either $Q\stackrel{\smash{\raisebox{-1ex}{\kern-0.75ex\scriptsize out}}}{}$ or $Q\stackrel{\smash{\raisebox{-1ex}{\kern-0.75ex\scriptsize in}}}{}$, thus reinforcing the validity of our methodology.
 As an alternative path, we leveraged the application of the argmax operation to the vector, $f(x)$, and finally attributed the logit scaled, the maximum probabilistic value obtained as ‘confidence’. This eliminates the need to know the actual labels of the data point, thereby freeing us from label-dependent constraints.
\begin{equation}
\text{confidence} = \log(\text{argmax}(f(x)))
\end{equation}
Subsequently, a parametric likelihood ratio test was employed as the conclusive step in our methodology to infer the membership of the target point. This Likelihood Ratio Test was used as it is proven more powerful at low false positive rates, making the attack more versatile.
}

\section{Results} {
In \cite{carlini2022membership}, the authors employed LOSS Values (baseline) for fitting IN and OUT models. In reference to Table 1, the first column shows the baseline, while the next four columns display our results. Our findings demonstrate that we have achieved AUC metrics that are commensurate with the anticipated performance levels.
  
As indicated above, our approach has led to successful results in various evaluation metrics. A good AUC shows that our methods excel in distinguishing between positive and negative classes across varying decision thresholds. Further, due to the lack of availability of computational resources, our results are restricted. They can additionally be improved to a great extent by scaling up the number of shadow models and increasing the epoch count.

}
\section{Future Work} {

Modifications to the baseline LiRA ratio can be explored to enhance our approach further. One avenue for improvement involves considering the difference between confidence values obtained for the IN and OUT models. Instead of employing logarithmic scaling for confidence values, we could further investigate other transformations that may help us model a Gaussian distribution from the confidence values. Our confidence methodology can also be further applied to other membership inference (MI) attacks. Owing to computational constraints, our model was trained for a limited duration of 21 epochs. Consequently, there exists the potential for further enhancements in results through the allocation of additional computational resources.

\begin{figure}

    \centering
    \includegraphics[scale=0.26]{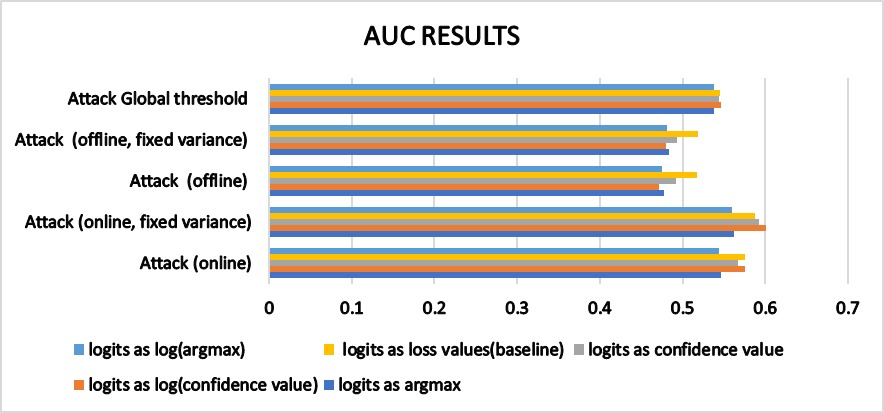}
    
    \caption: {Comparison of different methods using AUC as metric\newline (Color coding: Blue = Attack Global Threshold\newline Gold=Attack (offline,Fixed Variance)\newline Grey= Attack (Offline)\newline Orange = Attack (Online,Fixed Variance)\newline 
 Dark Blue= Attack (online) )}

    \label{fig:image_results}
\end{figure}
}

\section{Conclusion}{

The above table demonstrates that we have achieved results on par with those derived from the LiRA baseline approach. The assurance of our
 attack’s success remains steadfast, and our second methodology also allows us to carry out this attack without knowing the actual labels of data points, making the use of confidence values a viable metric in the case of MIA attacks. 

}

\bibliography{mybib.bib} 

\end{document}